# Transfer Learning for Estimation of Pendubot Angular Position Using Deep Neural Networks

Sina Khanagha
Electrical Engineering Department
Iran University of Science and Technology
Tehran, Iran
khanagha_s@elec.iust.ac.ir

*Abstract*— **In this paper, a machine learning based approach is introduced to estimate Pendubot angular position from its captured images. Initially, a baseline algorithm is introduced to estimate the angle using conventional image processing technique. The baseline algorithm performs well for the cases that the Pendubot is not moving fast. However, when moving quickly due to a free fall, the Pendubot appears as a blurred object in the captured image in a way that the baseline algorithm fails to estimate the angle. Consequently, a Deep Neural Network (DNN) based algorithm is introduced to cope with this challenge. The approach relies on the concept of transfer learning to allow the training of the DNN on a very small fine-tuning dataset. The base algorithm is used to create the ground truth labels of the fine-tuning dataset. Experimental results on the held-out evaluation set show that the proposed approach achieves a median absolute error of 0.02 and 0.06 degrees for the sharp and blurry images respectively.**

## I. Introduction

The Pendubot is an underactuated two-link robot that is widely used in the study of nonlinear control systems. The first joint of Pendubot (lower link) is actuated by a DC-motor and the second joint (upper link) is unactuated and moves freely around the first link (Fig. 1). The goal of the control is to balance the linkage in the unstable inverted equilibrium [1, 2]. To achieve and maintain the balance of the Pendubot, precise knowledge of the arms angular positions is essential. The angular position of the joints is usually provided by rotary encoders connected to the Pendubot. In this study, we consider the case that the rotary encoder is not available and investigate the possibility of using a basic camera to estimate the angles. In order to simplify this estimation, each one of the Pendubot arms are marked with a dark rectangle, as shown in Fig 2. We Initially investigate the use of conventional image processing techniques to detect the dark rectangle, and use its spatial location to estimate the angle. This approach performs well when the Pendubot is moving slowly and appears as a sharp object. However, the upper Pendubot arm sometimes falls freely with a very high velocity such that it appears as a blurred object in the captured image. Moreover, the quality of the camera, specifically the slow shutter speed of the cameras, may significantly increase the amount of motion blur in the image. As a result, the basic object detection algorithm fails in detecting the blurred Pendubot image.

Consequently, a Deep Neural Network (DNN) based machine learning approach is introduced to reliably estimate the required angles from blurry Pendubot images. The approach is base on the concept of transfer learning [3] to allow the training of the DNN with a relatively small labeled dataset of images. The well-known pre-trained VGG network [4] is used for this purpose and is fine-tuned on a small dataset of raw and blurry images. The labels for the dataset are obtained by applying the basic object detection algorithm on a set of sharp images that are then augmented to create a blurry version of the dataset for fine tuning. The experimental results on a held-out dataset shows that the proposed approach achieves reliable estimates of the Pendubot for the majority of sharp and blurred images.

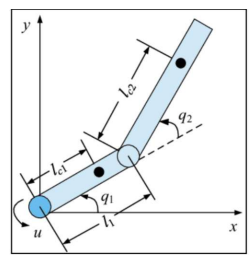

Figure 1 – Schematic diagram of a Pendubot.

This paper is organized as follows. In section II, the basic object detection algorithm is introduced. In section III, the Machine Learning approach is presented and the experimental results are reported in section IV, the experimental results are presented. Finally the conclusions are drawn in section V.

## II. The baseline Object Detection Approach

In this section, we introduce the heuristic approach that relies on conventional image processing techniques to estimate Pendubot angular position from the images captured by a camera. To simplify the task, a dark rectangle is attached to the Pendubot arms, as shown in Fig. 1-a. The task is then simplified to recognizing the vertices of the dark rectangle and use their coordinates to estimate the angular position. This can be accomplished by applying a contour detection algorithm followed by a shape approximation algorithm.

Initially, a number of pre-processing functions are applied to the image to reduce the noise and to remove irrelevant patterns. First, the colored image is converted to a grayscale image, and followed by a binary threshold filter to create a binary black and white image. Given that our object of interest is inherently dark, applying an appropriate threshold plays an important role in removing many unwanted brighter patterns in the image. Fig. 1-a shows an example image captured by the camera and Fig. 1-b shows the resulting pre-processed image. In the final pre-processing step, Canny edge detection algorithm is applied to further remove the scattered noisy patterns and to reveal the borders of prominent shapes, specially the large rectangle of interest. Fig 1-c shows the final pre-processed image after the application of the edge detection filter. The image is now prepared for the contour detection algorithm.

A contour is an outline that represents or bounds the shape of an object. Contour detection is an especially useful tool in object detection, shape analysis and object recognition. We use Suzuki's Contour tracing algorithm [5] for structural analysis of digitized binary images to find all of the contours that exist in the pre-processed image. The dark rectangle of interest will be one of these contours, represented by all of the points on its boundary. In the next step, the Douglas-Peucker algorithm [6] is applied to approximate each contour with a new contour made of the smallest set of points such that the distance between the two contours is smaller than a threshold. As a result, each polygon in the image will be represented by a compressed contours that only includes the corresponding vertices. For instance, the desired dark rectangle would be represented as a contour only made of its four vertices. It can then be distinguished from other detected contours using heuristic rules on the number of detected vertices and the area of the detected object. For instance, any contour made of four points is a rectangle, and the rectangle of interest can be pinned down using manually set threshold on the area it covers.

Once the vertices of the coordinates of the dark rectangle are detected, their coordinates are used to estimate the angular position of the Pendubot. First, the coordinates of the nearby vertices are averaged to find the middle of the top and bottom sides of the rectangle whose coordinates can be shown as $(x_1, y_1)$ and $(x_2, y_2)$. These coordinates are used to estimate the angle $\theta$ between a line connecting these two points with respect to the horizontal axis:

$$\theta = \arctan\left(\frac{y_1 - y_2}{x_1 - x_2}\right) \quad (1)$$

Fig. 1-d shows an example of the detected rectangle (green lines connecting the four detected vertices), the line connecting the middles of the two smaller sides and the estimated angle of the Pendubot.

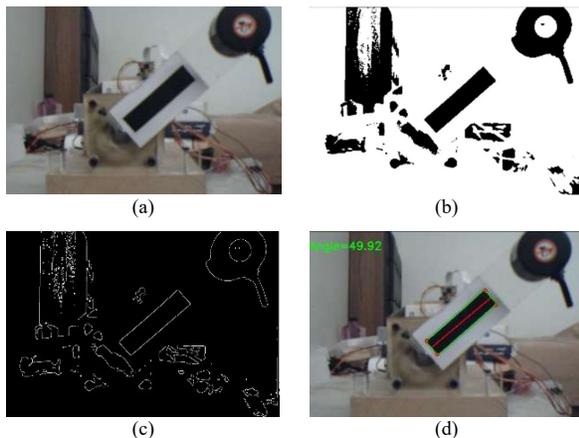

Figure 2 – (a) lower arm of a Pendubot marked with a dark rectangle to simplify its detection, (b) the pre-processed black and white image, (c) applying edge detection reveals prominent shapes, (d) dark rectangle is detected with its vertices shown and the angle of the line connecting the middle of its two smaller sides is estimated w.r.t the horizontal axis.

This algorithm can be easily hand tuned for different camera/lighting setups and can reliably estimate the angle as long as the dark rectangle is captured as a sharp object in the image. However, in the case that the Pendubot moves with a high velocity due to a free fall, the dark rectangle becomes blurry and given that its sharp corners and vertices are no longer clearly captures, they can no longer be found by the Douglas-Peucker algorithm. In the next section, a machine learning solution is introduced to cope with the challenge of blurry images.

III. THE DEEP NEURAL NETWORK BASED APPROACH

As discussed in section II, the heuristic baseline algorithm performs well for sharp images, but not when the images are blurred. In this section we introduce a machine learning based approach to estimate Pendubot angular position in such challenging cases. Specifically, we consider Deep Neural Network (DNN) based approaches that have shown great success in complex computer vision problems [7, 8]. A significant challenge in the use of deep neural networks for a new applications is to collect a reasonably large training dataset to be used for the training of the DNN, so that it learns the patterns of interest. In the case of estimating Pendubot angular estimation, the goal is to reliably estimate the angular position of a potentially blurred dark rectangle, regardless of Pendubots background, which may include any real-life visual context in a variety of lighting conditions. The collection of a large dataset that fairly represents all such visual contexts is indeed a challenging and time consuming task.

In such a case, an alternative approach is to use transfer learning where the weights of a DNN trained on a different but similar task associated with a huge public benchmark dataset, are used as a starting point and then, the pre-trained DNN is fine-tuned on the desired task using a significantly smaller training dataset. As such, the knowledge learnt by the pre-trained DNN is transferred and reused for the new machine learning task, while significantly reducing DNN's training time and its risk of being overfit to a small training dataset [11]. Depending on the target task, one can either use the pre-trained network only to initialize the weights and fine-tune all the weights or to freeze a number of layers, and fine-tune only the last few layers of the pre-trained DNN.

The architecture we use for deep transfer learning is the well-known VGG network that is composed of several layers of Convolutional Neural Networks (CNN) followed by 3 layers of feed forward neural networks that perform the image classification task [4]. VGG is trained on the ImageNet dataset consisting of more than 14 million images belonging to nearly 1000 classes [9]. In the case of estimating the angular position of a Pendubot, the objective is to estimate a continuous value (regression) rather than predicting the class each image belongs to (classification). We thus remove the three feedforward layers of VGG that conduct the image classification task and replace them with four feedforward layers that perform a regression task. The size of feedforward layers are set to 128, 64, 64 and 64. The activation function used for these layers include Rectified Linear Units (Relu) for the first three layers and a Sigmoid for the final layer. A Mean Squared Error loss function is used to perform fine-tuning for these three layers. The CNN layers of the VGG are frozen and kept unchanged during the fine-tuning process.

## IV. EXPERIMENTAL RESULTS

Although the use of transfer learning reduces the need for labeled dataset, we still need to collect a reasonable amount of label data to conduct fine-tuning and to evaluate the overall performance of the system. We use the baseline algorithm introduced in section 2 to create a small fine-tuning dataset. However, given that the baseline algorithm cannot estimate the angle from blurry images we can only obtain labels for sharp images. In order to obtain labeled blurry images for fine-tuning, a gaussian kernel is applied to the collected sharp images that are labeled by the baseline algorithm. This is a common technique in the field of computer vision called data augmentation [10]. Other augmentation techniques such as channel and brightness shift and vertical and horizontal shift are also randomly applied to further diversify the training dataset.

We collected four batches of images, in four different lighting and background conditions. Each batch contains at least 20 images for all integer Pendubot angles between -90 to 90 degrees. All these batches are augmented to obtain their blurry version too. We set aside one of the original batches, along with its augmented blurred version, as our unseen evaluation set and use the remaining six batches for fine tuning. Table 1 summarizes the number of images for the training and validation sets. A total of 22886 raw images are used for training and 3963 raw images, captured in a completely different camera setup are held out as the unseen test dataset. Fig. 3 shows two examples of raw and blurry images from the evaluation dataset.

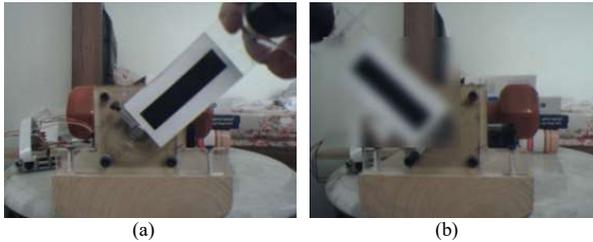

(a) (b)

Figure 3 - (a) a raw test image, (b) an augmented test image to introduce blurriness to the raw image.

The performance of the angle estimator is evaluated in terms of Absolute Estimation Error (AEE). The results are presented in Table I. It can be seen that the estimation errors are very small: the median value is close to zero for both raw and blurry images and 90% of the errors are smaller than 2.87. It is important to note that the results were obtained assuming a generic estimator with variable background and flexible camera distance and location. It is possible to significantly reduce the error by creating a fixed setting, such that the images have a fixed background and the distance between the camera and Pendubot is constant.

We finally compared the computational complexity of the DNN based approach with that of the baseline algorithm. Comparing the run time for estimating the angle for all of the raw images in the evaluation set on a same computer, showed that the DNN based approach is only 4.88 times slower than the baseline approach. Given that the baseline approach completely fails in estimating the angle from the blurred images, the slight increase in processing time can be acceptable as long as the algorithm can be run in real-time.

TABLE I. PERFORMANCE RESULTS IN TERMS OF ABSOLUTE ESTIMATION ERROR (AEE).

| Dataset | Raw Images | Blurry Images | All images |
|---|---|---|---|
| Number of images | 3963 | 3963 | 7926 |
| Mean AEE | 1.52 | 2.02 | 1.77 |
| Median AEE | 0.02 | 0.09 | 0.04 |
| Standard deviation of AEE | 1.29 | 1.72 | 1.52 |
| 90th percentile of AEE | 2.47 | 3.27 | 2.87 |

## V. CONCLUSIONS

In this paper, we proposed a machine learning based approach to estimate the angular position of a Pendubot from the images captured by a camera. The approach was developed using the concept of transfer learning to train a large DNN using a small fine-tuning dataset. The labels for the dataset was created using a baseline algorithm developed by conventional image processing techniques. We showed that the DNN based approach achieves reasonable accuracy in estimation of the angular position from sharp images. More importantly, it can accurately estimate the angle from images that are blurred due to the movements of the Pendubot, for which the baseline algorithm completely fails to make an estimation. Moreover, the proposed DNN based algorithm performed well in a completely new lighting and camera position setting, while the baseline algorithm requires manual setting of thresholds and other hyper parameters for each new camera setup. This shows that the proposed method provides a reliable and flexible solution for the estimation problem studied in the paper.